\documentclass{Interspeech2024}

\interspeechcameraready

\title{Naturalistic Language-related Movie-Watching fMRI Task for Detecting Neurocognitive Decline and Disorder}
\usepackage{makecell}
\usepackage{bm}
\usepackage{siunitx}
\usepackage{multirow}
\usepackage{amsmath}
\usepackage{cite}

\name[affiliation={1*}]{Yuejiao}{Wang}
\name[affiliation={1*}]{Xianmin}{Gong}

\name[affiliation={1}]{Xixin}{Wu}
\name[affiliation={1}]{Patrick}{Wong}
\name[affiliation={1}]{Hoi-lam Helene}{Fung}

\name[affiliation={2}]{Man Wai}{Mak}
\name[affiliation={1}]{Helen}{Meng}

\address{
  $^1$The Chinese University of Hong Kong
  $^2$The Hong Kong Polytechnic University
 }

\email{\{wangy, wuxx, hmmeng\}@se.cuhk.edu.hk {xianmingong, p.wong, hhlfung}@cuhk.edu.hk man.wai.mak@polyu.edu.hk $^*$Equal contributions}


\begin{document}

\maketitle

\begin{abstract}
Early detection is crucial for timely intervention aimed at preventing and slowing the progression of neurocognitive disorder (NCD), a common and significant health problem among the aging population. Recent evidence has suggested that language-related functional magnetic resonance imaging (fMRI) may be a promising approach for detecting cognitive decline and early NCD. In this paper, we proposed a novel, naturalistic language-related fMRI task for this purpose.  We examined the effectiveness of this task among 97 non-demented Chinese older adults from Hong Kong. The results showed that machine-learning classification models based on fMRI features extracted from the task and demographics (age, gender, and education year) achieved an average area under the curve of 0.86 when classifying participants' cognitive status (labeled as NORMAL vs DECLINE based on their scores on a standard neurcognitive test). Feature localization revealed that the fMRI features most frequently selected by the data-driven approach came primarily from brain regions associated with language processing, such as the superior temporal gyrus, middle temporal gyrus, and right cerebellum. The study demonstrated the potential of the naturalistic language-related fMRI task for early detection of aging-related cognitive decline and NCD.
 
\end{abstract}

\section{Introduction}
The growing aging population poses a significant challenge to society. In China, older adults aged 60 and above constitute 18.70\% of the total population \cite{Census2023}. Aging typically leads to neurocognitive decline, which may progress into neurocognitive disorder (NCD) that is more severe and goes beyond the normal cognitive aging trajectory. Globally, over 55 million people live with dementia (i.e., major NCD, a late stage of NCD) \cite{dementia2023}, with approximately 15 million cases in China \cite{elderlyreport2021}. NCD significantly undermines individuals’ daily functioning and well-being and places a heavy burden on society. Therefore, early detection of cognitive decline and NCD is crucial, enabling timely intervention to prevent or slow down NCD progression.

Multiple approaches have been proposed for early NCD detection, but none has been entirely satisfactory so far \cite{alzheimer_diagnosis_2021,venugopalan_multimodal_2021,pan_early_2020}. Recent evidence suggests that examining brain language function via language tasks combined with neuroimaging (e.g., functional magnetic resonance imaging (fMRI) and electroencephalogram) is a promising approach. Degradation in brain language function is common in various types of NCD, and such changes may appear early before noticeable NCD symptoms emerge \cite{reilly2010cognition,jarrold2014aided}. However, existing language tasks (e.g., the classic picture-naming task, which requires participants to speak out the names of a series of objects) used for screening and detecting NCD are typically simple and focus on basic language functions at the word or verbal level \cite{cullen2007review,lonie2009screening}. Additionally, most of these tasks are laboratory-based and lack ecological validity. These tasks differ from daily activities and may not accurately reflect people’s cognitive functioning in real life \cite{law2012measures,bielak2017cognitive}. These drawbacks may limit the effectiveness of the existing language-related fMRI tasks in detecting NCD.

\begin{figure}[t]
  
  \centering
  \includegraphics[width=\linewidth]{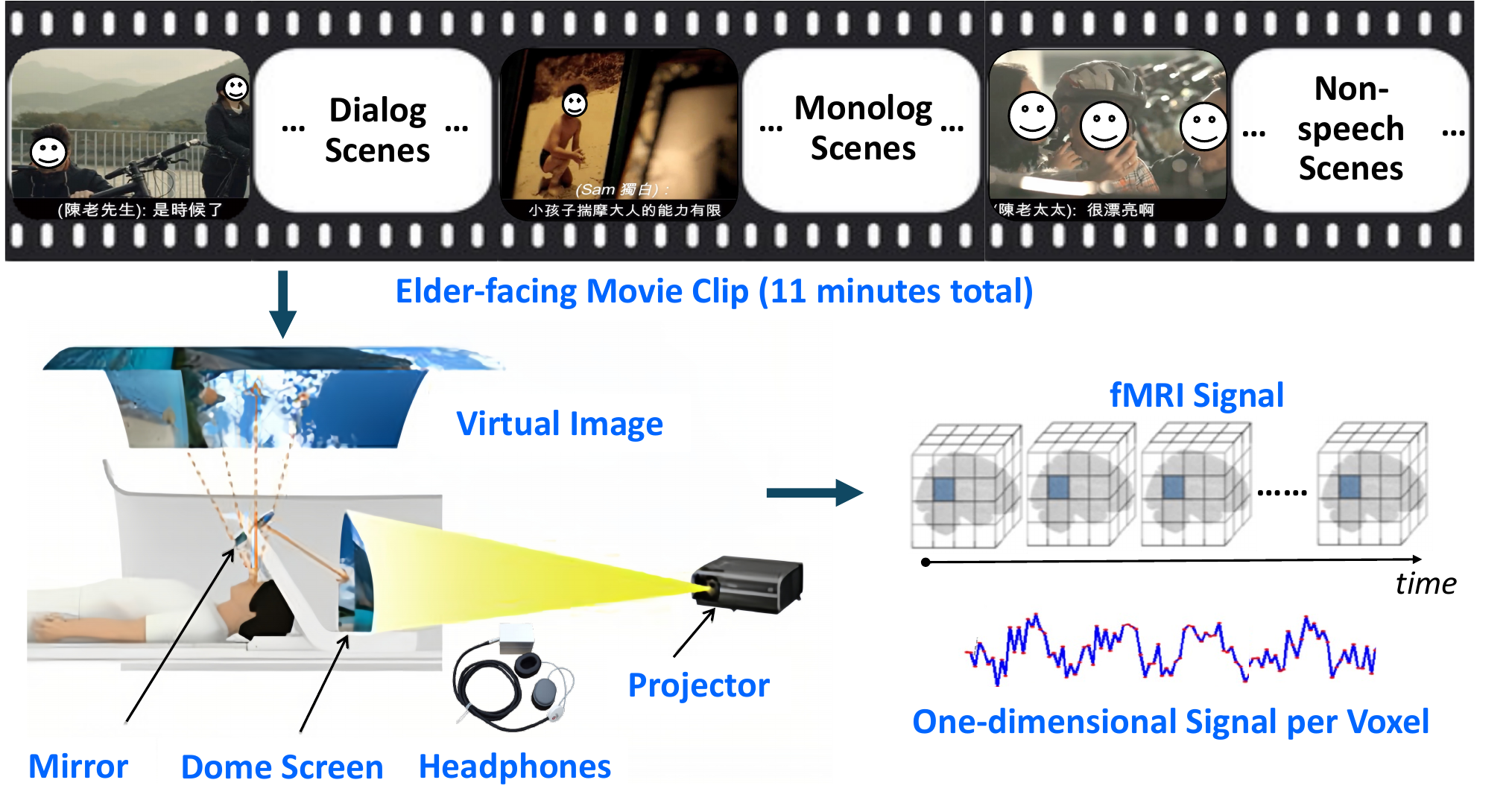}
  \vspace{-1em}
  \caption{The proposed movie-watching fMRI task. While being scanned in the MRI scanner, the participant watches a movie projected onto a mirror placed inside the scanner. The participant receives the movie's audio stream through headphones.}
  
  \label{task_plot}
  \vspace{-2em}
\end{figure}

In this paper, we introduce a novel approach for detecting cognitive decline and NCD, which integrates a naturalistic language-related fMRI task, a movie-watching task, and functional magnetic resonance imaging. This approach possesses several advantages. First, the movie-watching task closely resembles a common daily activity (i.e., watching movies), making it more realistic and ecologically valid than traditional laboratory-based tasks. Studies have shown that naturalistic tasks are better at detecting cognitive decline and impairment \cite{law2012measures,bielak2017cognitive}. Second, the task involves multiple levels of language functions, ranging from functions processing phonetic and articulatory linguistic features at the lower level to functions processing discourse and pragmatic linguistic features at the higher level. Those higher levels of language functions, not well captured in the traditional laboratory-based tasks, are more susceptible to NCD and could be more sensitive for detecting NCD \cite{aggio2018cognition,kempler2008language}. Third, the task only requires participants to passively watch a short movie clip without actively performing operations. This makes the task easy and friendly even for participants with difficulty comprehending or following instructions. Fourth, the language functions elicited by the task are measured using fMRI, which is non-invasive and provides high spatial resolution. The fMRI technology has been widely used and proven to be powerful for studying brain language functions \cite{bookheimer2002functional,price2012review}. The measure does not rely on participants’ self-report and assessors’ judgments, providing a more objective and sensitive measure of language functions. 

In the following section, we will demonstrate the effectiveness of this approach in detecting cognitive decline (before the development of dementia or major NCD) among Hong Kong Chinese. Specifically, we scanned Hong Kong Chinese older adults while they were watching a short movie in Cantonese. We then trained and cross-validated models based on the fMRI data to classify these older adults into a binary group membership (cognitively NORMAL vs. DECLINE) defined by their scores on the Montreal Cognitive Assessment (MoCA) \cite{nasreddine2005montreal}, a well-established test for assessing cognitive function and assisting NCD diagnosis. This study supports the potential of the movie-watching task to detect cognitive decline before the appearance of apparent NCD symptoms.

\begin{table}[tbp]
\caption{Statistics of demographics and MoCA scores of participants in the Movie-watching fMRI task and MoCA test.}
\vspace{-0.5em}
  \label{tab: Dataset}
  \centering
  \renewcommand\arraystretch{1.1}
\begin{tabular}{ccccc}
\toprule
\multirow{2}{*}{\textbf{Feature}} & \multicolumn{2}{c}{\textbf{Male} (n = 54)} & \multicolumn{2}{c}{\textbf{Female} (n = 43)} \\ \cline{2-5} 
                                & Mean             & Std.             & Mean               & Std.               \\ \midrule
Age                               & 72.02                & 5.78            & 70.74                  & 6.28               \\
Education                               & 9.02                & 3.71            & 6.65                  & 3.62              \\
MoCA                              &  21.00               & 4.08            & 19.12                  & 4.19               \\ 
\bottomrule
\vspace{-3em}
\end{tabular}
\end{table}

\section{Data Collection}

\textbf{Participants.} 97 Chinese older adults aged above 60 years (see Table~\ref{tab: Dataset} for more demographic information) residing in Hong Kong were recruited. All participants were right-handed, had normal or correct-normal visual acuity, could read Chinese, without dementia, and were eligible for fMRI scanning. Ethical approval was obtained from the medical ethics committee. Informed consent was obtained from all participants.

\textbf{The movie-watching task.} As shown in Figure~\ref{task_plot}, participants watched a short Cantonese movie (containing visual and auditory streams and subtitles), called “Sweet Home”, while lying still in an MRI scanner. The movie lasted for 662 seconds and depicted daily interactions and conversations within a family. The scenarios in the movie closely resembled typical daily situations in Hong Kong families and were meant to be familiar to participants. Some movie segments contained speeches (dialogues or monologues), some were silent, and these segments were interspersed throughout the movie. To motivate participants to engage with the movie, they were informed beforehand that they would be asked questions about the movie afterward.

\textbf{FMRI data acquisition and preprocessing.} FMRI data were acquired when participants were watching the movie in the MRI scanner. T2-weighted blood-oxygen-level-dependent signals were collected using a Siemens MAGNETOM Prisma 3 Tesla MRI Scanner with a 64-Channel Head/Neck coil. Multiband (factor = 6) gradient echo planar sequence was used to scan the whole brain. The scanning parameters were repetition time = 900 ms, echo time = 24 ms, flip angle = 90°, voxel size = 2 × 2 × 2 mm\textsuperscript{3}, matrix size = 104 × 104, field of view = 206 × 206 mm\textsuperscript{2}, and number of slices = 72. The first nine fMRI volumes were discarded to ensure magnetization stabilization, leaving 736 volumes per subject for analysis. We used SPM12 (fil.ion.ucl.ac.uk/spm/software/spm12) to preprocess the functional images. Following the standardize protocols \cite{della2002empirical}, the preprocessing included field map correction, realignment, slice timing correction, co-registration, segmentation, normalization to the standard Montreal Neurological Institute (MNI) space, and spatial smoothing with an isotropic 5 mm full-width-at-half-maximum Gaussian kernel. 

\textbf{MoCA test for cognitive status.} As this study aims to examine whether speech-related fMRI features derived from the movie-watching task is effective in detecting cognitive decline before overt NCD symptoms emerge (and thus can be used for early NCD detection), all participants recruited in the present study did not meet the diagnostic criteria of major NCD (or dementia). Their cognitive status was measured using the Hong Kong version of MoCA \cite{wong2009validity}. The test score ranged between 0 and 30, with a higher score indicating a better cognitive status. We categorized participants into two groups for classification based on their MoCA scores (see details in section \ref{class_task}).

\section{Approach}

\begin{figure}[t]
  
  \centering
  \includegraphics[width=\linewidth]{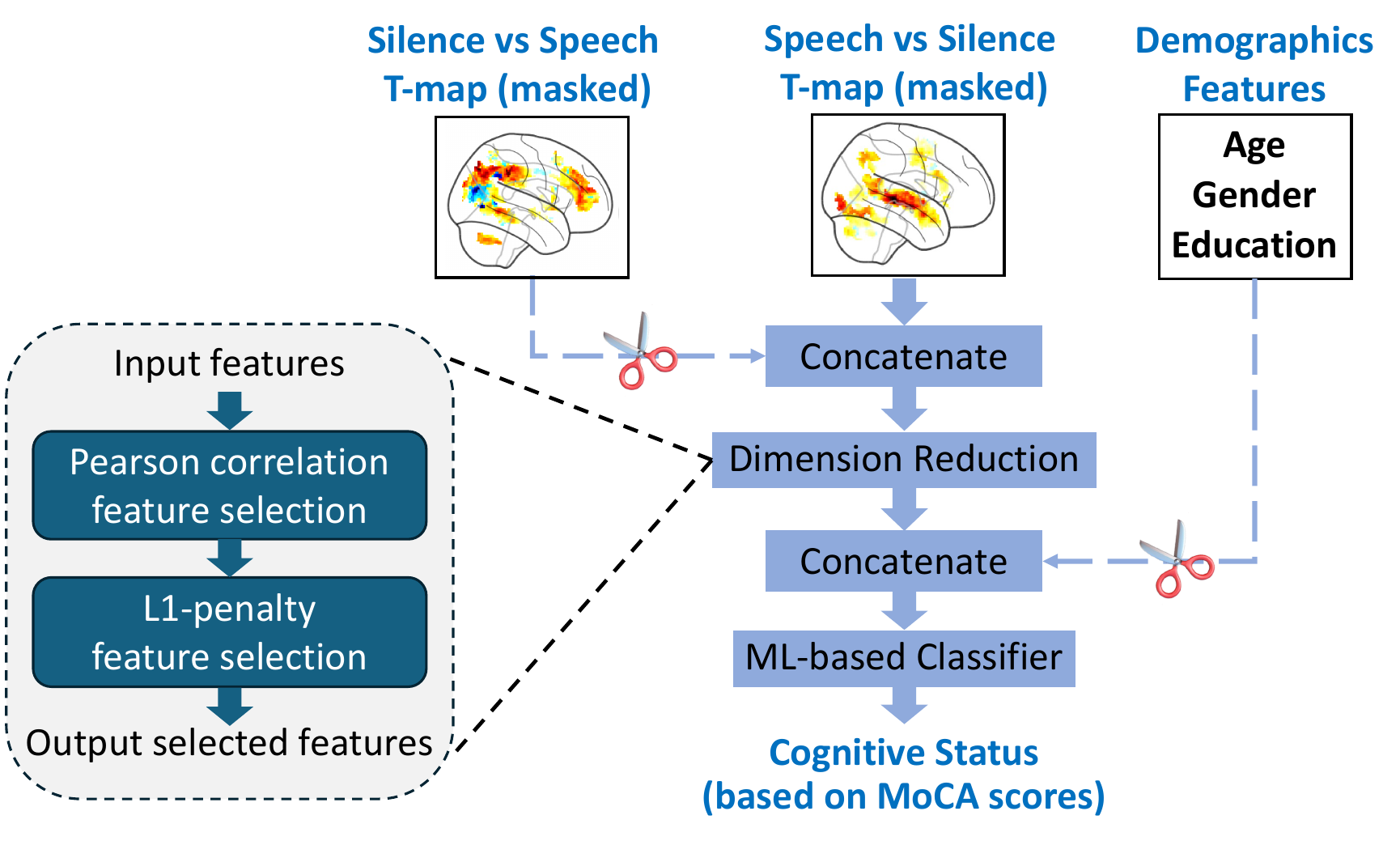}
  \caption{The flowchart of machine learning-based cognitive status classification. T-maps obtained from the movie-watching fMRI task underwent a two-step dimension reduction process and were concatenated with demographics for cognitive status classification. Scissors means the feature is optional.}
  \label{ML_flowchart}
  \vspace{-1.5em}
\end{figure}

\subsection{Language-related Statistical T-maps Extraction}

For each participant, we used general linear modeling to analyze their brain activation related to speech processing:
\vspace{-0.5em}
\begin{align}
\label{tmap_equ}
    Y=\beta{_0}+\beta{_1}(Speech \,event * HRF) \notag \\ + \beta{_2}(Silence \,event * HRF) \notag \\ +Headmotion \, Regressors +Residual 
\end{align}
where the fMRI signals Y (a 1-D time series) within each brain voxel of each participant was regressed on two event vectors that encoded the presence of speech and silence at different time points during the movie: 1) a speech-event vector (0 = absence of speech, 1 = presence of speech); and 2) a silence-event vector (0 = non-silence, 1 = presence of silence). Both event vectors were convolved with the canonical Hemodynamic Response Function (HRF). The HRF models the time-dependent change in fMRI signals after a specific neural activity is evoked by an event (e.g., the presence of speech in this study) \cite{buxton2004modeling}. The resulting regression coefficient $\beta{_1}$ and $\beta{_2}$ indicated the respective contributions of the speech-event vector and the silence-event vector to the fMRI signals within a specific brain voxel. 

Since a speech event represented a movie segment containing both visual and speech information, while a silence event represented a movie segment containing only visual information, the contrast between $\beta_1$ and $\beta_2$ reflected the isolated effect of speech on the fMRI signals. We constructed two contrasts, $C{_1}$ = $\beta_1$ - $\beta_2$ and $C{_2}$ = $\beta_2$ - $\beta_1$, and kept only positive values for each. A larger $C{_1}$ value indicated that a voxel was more strongly activated by speech, while a larger $C{_2}$ indicated that it was more strongly activated by silence or deactivated by speech. To account for the varying estimation uncertainty, each brain voxel’s contrasts $C{_1}$ and $C{_2}$ were further transformed to t-statistic values (T${_1}$ and T${_2}$) by applying the formula T${_i}$= $C{_i}$ / pooled standard error of $\beta_1$ and $\beta_2$. The T values could be further used for statistical significance testing and developing classification models \cite{bron2015feature}. Eventually, two whole-brain T-maps were generated for each participant, with one containing T values indicating each brain voxel's activation by the presence of speech (i.e., Speech vs Silence T map) and the other containing T values indicating deactivation by the presence of speech (i.e., Silence vs Speech T map).

Note that the regression model above also included head-motion parameters as additional regressors. The purpose was to regress out the undesirable influences of head motion on the fMRI signals \cite{power2012spurious}. Additionally, a high-pass filter with a cutoff period of 128s was applied to the fMRI signals to remove low-frequency signals, and a first-order autoregression AR(1) was used to control temporal autocorrelations. 

\subsection{Detection of Cognitive Decline}
\label{class_task}
Language-related brain functions could change with cognitive decline even before the emergence of overt NCD symptoms. Therefore, we used the speech-related brain T-maps, together with demographics that could influence cognitive ability (including age, gender, and education year), as features to classify participants' cognitive status.

\begin{figure*}[htb]
  \centering
  \includegraphics[width=\textwidth]{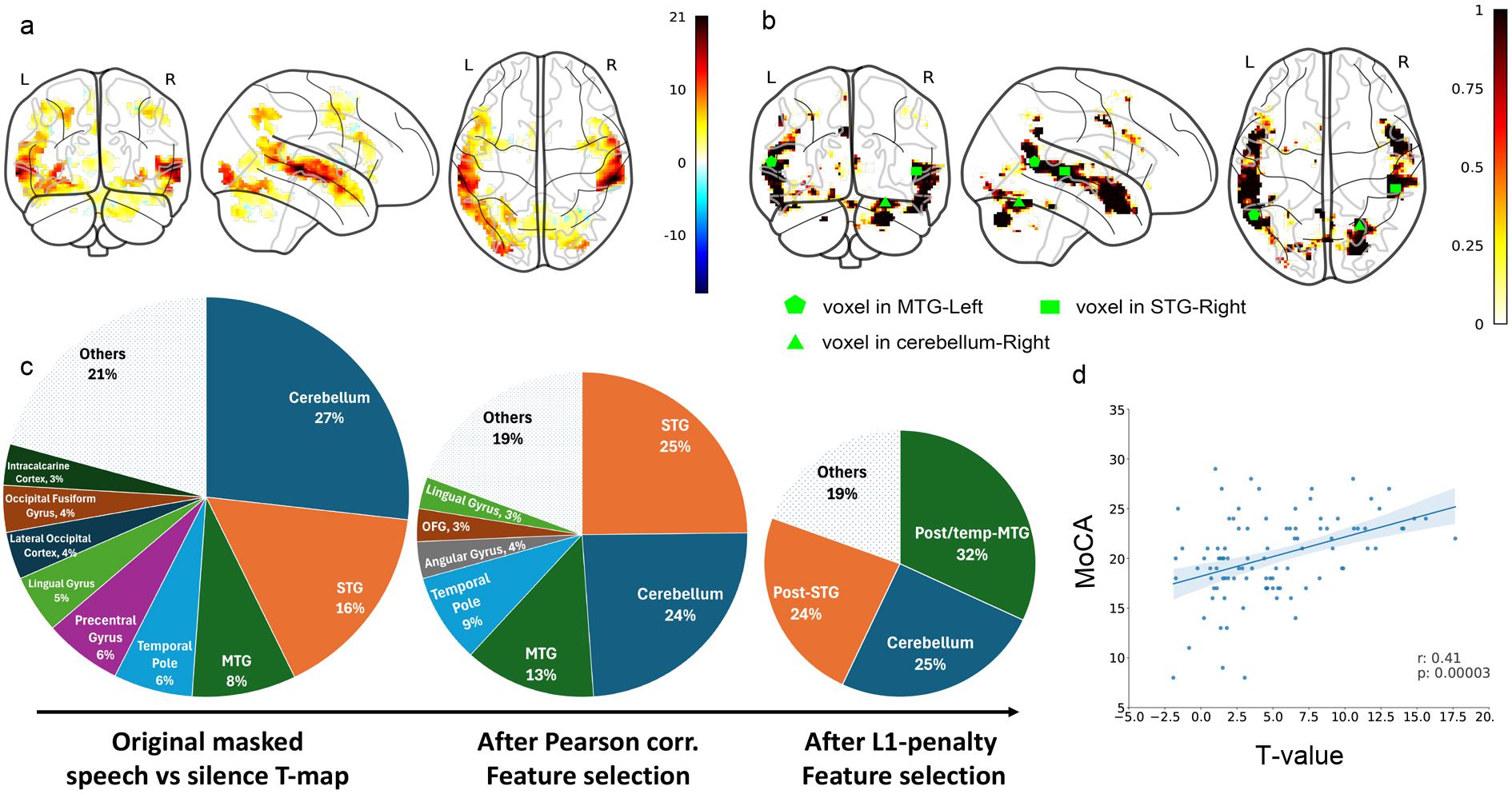}
  \vspace{-1em}
  \caption{(a) An example Speech vs Silence T-map in three different view perspectives (from a female subject with age = 63 and MoCA score = 24). (b) The probabilistic map showing the probability (out of 500 times of experiments) of each voxel being selected after the step-1 feature selection, with green markers representing the top three most frequently selected voxel clusters after the step-2 feature selection. (c) The distributions of voxels across brain regions before and after feature selections. (d) The scatter plot of T values from the right STG and MoCA scores of 97 participants. L and R in (a) and (b) means the left and right hemispheres. MTG: middle temporal gyrus. STG: superior temporal gyrus. OFG: occipital fusiform gyrus.}
  \label{big_fig}
  \vspace{-1.5em}
\end{figure*}

\textbf{ML-based classification.} According to subjects' MoCA scores (median= 20), we labelled them as the cognitive NORMAL ($MoCA>20$) or DECLINE ($MoCA\leq20$) group.
The training flowchart of ML-based classification is shown in Figure~\ref{ML_flowchart}. For the brain T maps (i.e., Speech vs Silence T maps and Silence vs Speech T maps), we only included brain voxels as features if their $C{_i}$ values were significantly greater than 0 (i.e., T values above the statistical significance threshold) across all participants. We term these T maps as \textbf{masked T maps}. To further reduce the dimension of features, we applied a two-step feature selection procedure to the masked T maps. 1) Firstly, we kept only those voxels with T values significantly correlated with MoCA scores across all training participants (\textit{p} of Pearson correlation $<$ 0.01). 2) Secondly, we applied a L1-penalty feature selection. Specifically, we used the \textit{SelectFromModel} function of the \textit{sklearn} toolkit, with the default threshold of 1e-5. This function selected features based on the importance weights from a linear Support Vector Classifier (SVC) model with L1 regularization penalty. 

In this way, voxels that were redundant or contributed little to the classification task were removed from the feature set. 
Demographics features could also influence cognitive function \cite{rajan2021population}. They were also included as optional features without feature selection.
Finally, features were fed into an SVC (C = 1.0 as default) or Gaussian Naive Bayes (GNB) model for classifying participants into the binary cognitive status labels.

\textbf{Experimental design.}
To evaluate the effectiveness of the classification models, we conducted 500 iterations of stratified shuffle-split cross-validation using the \textit{StratifiedShuffleSplit} function of the \textit{sklearn} toolkit. In each iteration of cross-validation, participants were split into a training set (95\% of the participants) and test set (5\% of the participants). The two-step feature reduction process (as described above) and the final classifiers were fitted on the training set and then applied on the test set. The performance was assessed using the area under the curve (AUC) metric on the test set. The 500 iterations of cross-validation resulted in an AUC distribution with 500 samples. 

We compared the performance of different feature combinations: 1) demographics only; 2) Speech vs Silence T-map feature only; 3) Speech vs Silence T-map and Silence vs Speech T-map features; 4) demographics and Speech vs Silence T-map features; and 5) all three types of features. To determine whether a feature combination performed significantly better than another, we conducted a Wilcoxon signed-rank test \cite{woolson2005wilcoxon} on their AUC distributions (a nonparametric statistical tests), using the \textit{wilcoxon} function in \textit{scipy} toolkit.

\section{Results and Discussions}

\subsection{T-maps' Coverage of Language-related Brain Areas }
To demonstrate that the Speech vs Silence T-maps extracted from the movie-watching task (see Figure~\ref{big_fig}a) captured the activity of brain regions associated with language processing, we compared the distribution of voxels in the masked Speech vs Silence T-map against a validated probabilistic functional atlas of the language network (\textit{LanA}) \cite{lipkin2022probabilistic}. 

At the voxel level, the voxels within the masked Speech vs Silence T-map overlapped with 78.25\% of voxels in \textit{LanA} (with $P>0.1$). 
At the brain region level (using labels from the Harvard-Oxford atlas), voxels with $P>0.5$ in \textit{LanA} are located within 17 brain regions, including 25\% of voxels within the superior temporal gyrus (STG), 29.9\% within the middle temporal gyrus (MTG), and 10.7\% within the temporal pole. The voxels within the masked Speech vs Silence T-map were also mainly distributed within these three brain regions (refer to Figure~\ref{big_fig}c). These results proved that 1) the movie-watching task effectively activated the language-related brain regions, and 2) the Speech vs Silence T-map captured brain activity associated with language processing.

\subsection{Results of Cognitive Status Classification}

\begin{table}[t]
\caption{Performance (assessed using AUC) of models based on different feature combinations for cognitive status classification. AUC: area under the curve; SVC: support vector classifier; GNB: Gaussian naive Bayes.}
\vspace{-0.4em}
  \label{tab:AUC}
  \renewcommand\arraystretch{1.3}
  \centering
\begin{tabular}{lcccc}
\toprule

\multirow{2}{*}{\makecell[l]{Feature \\ Combinations}} & \multicolumn{2}{c}{SVC} & \multicolumn{2}{c}{GNB} \\ \cline{2-5} 

                                & Mean             & Std.             & Mean               & Std.                \\ \midrule
Demogra.                       & 0.730                & 0.252           & 0.755                  & 0.246       \\
Speech vs silence T-map        & 0.810                & 0.212            & 0.799                  & 0.214       \\
\makecell[l]{Demogra. with \\speech vs silence T-map} & 0.828                & 0.200    & 0.822       & 0.213      \\
Two T-maps                    & 0.846   & 0.200          & 0.834                 & 0.199           \\
All features                           & \textbf{0.862}              & 0.178          & \textbf{0.852}                  & 0.195   \\ \bottomrule
   
\end{tabular}
\vspace{-2em}
\end{table}

The average AUC of 500 stratified shuffle-split iterations is shown in Table~\ref{tab:AUC}. As expected, the performance of models based on demographic features was not bad, due to the influence of age and education on cognitive abilities. The models based on the Speech vs Silence T-map features were significantly better than that based on demographic features (Wilcoxon test $p<0.005$). When combining demographics with Speech vs Silence T-map features, the averaged AUC increased to above 0.82, which was significantly higher than models using only one type of features ($p<0.005$). These results indicated that the T-map features additionally contributed to the detection of cognitive decline on top of demographics.

Furthermore, the models combined with Silence vs Speech T-map features (reflecting deactivation by the presence of speech) also performed better than the demographics-only models ($p<0.05$). The Silence vs Speech T-map covered brain regions such as the posterior cingulate cortex and precuneus cortex, which largely overlap with the Default Mode Network in the brain that tends to be deactivated when a person is more strongly engaged in cognitive tasks \cite{raichle2015brain}. The models based on all three types features achieved an average AUC above 0.85, outperforming the models based on one or two types of features ($p<0.05$). These results suggest the benefits of integrating demographics and language-related fMRI features (from brain regions activated and deactivated by speech) for detecting cognitive decline and early NCD. 

\vspace{-1em}
\subsection{Localization of Brain Areas}

In each stratified split, only voxels significantly correlated with MoCA score (step-1 feature selection) and contributed to cognitive status classification (step-2 feature selection) were selected. 
Across the 500 stratified shuffle-split iterations, the most frequently selected voxels were located in the MTG, STG, and cerebellum (see Figure~\ref{big_fig}b,c), and the top three were from the left MTG, right STG and right cerebellum (see the green dots in Figure~\ref{big_fig}b). The MTG \cite{davey2016exploring} and STG (including the Wernicke's language area \cite{binder2017current}) play a central role in language function, particularly language perception. The cerebellum has also been shown to engage in various language processes, such as listening comprehension and conversation tracking \cite{de2007cerebellum,murdoch2010cerebellum}.

These results indicate that the classification models were relying on language-related brain function features to distinguish cognitive status. This localization increases the interpretability of the models. Overall, this study demonstrates the utility of language-related fMRI features derived from the naturalistic language-related movie-watching task in detecting cognitive decline and in potentially detecting early NCD. 

\vspace{-0.5em}
\section{Conclusions}
In this paper, we introduced and demonstrated the effectiveness of a naturalistic language-related movie-watching fMRI task for detecting cognitive decline and early NCD. The language-related fMRI features (statistical T-maps) extracted from the task, combined with demographics, achieved an averaged AUC of 0.86 for classifying participants' cognitive status. The data-driven feature selection process identified the most useful fMRI features  as coming from brain areas (the MTG, STG, and cerebellum) tightly relevant to language processing.
A limitation of this study was that it focused solely on basic language perception, without examining which sub-processes (e.g., semantic processing \cite{verma2012semantic}, syntactic processing \cite{bickel2000syntactic}) of language perception was most useful for detecting cognitive status and NCD. Future studies should address this issue, as well as expand the investigation to include language production \cite{martin1983word}. Additionally, future studies could examine whether the approach proposed in this study is truly useful for classifying clinical NCD diagnoses. 

\vspace{-0.5em}
\section{Acknowledgements} 
This research is partially supported by the HKSARG Research Grants Council’s Theme-based Research Grant Scheme (Project No. T45-407/19N).

\bibliographystyle{IEEEtran}
\bibliography{mybib}

\begin{thebibliography}{10}
\providecommand{\url}[1]{#1}
\csname url@samestyle\endcsname
\providecommand{\newblock}{\relax}
\providecommand{\bibinfo}[2]{#2}
\providecommand{\BIBentrySTDinterwordspacing}{\spaceskip=0pt\relax}
\providecommand{\BIBentryALTinterwordstretchfactor}{4}
\providecommand{\BIBentryALTinterwordspacing}{\spaceskip=\fontdimen2\font plus
\BIBentryALTinterwordstretchfactor\fontdimen3\font minus \fontdimen4\font\relax}
\providecommand{\BIBforeignlanguage}[2]{{%
\expandafter\ifx\csname l@#1\endcsname\relax
\typeout{** WARNING: IEEEtran.bst: No hyphenation pattern has been}%
\typeout{** loaded for the language `#1'. Using the pattern for}%
\typeout{** the default language instead.}%
\else
\language=\csname l@#1\endcsname
\fi
#2}}
\providecommand{\BIBdecl}{\relax}
\BIBdecl

\bibitem{Census2023}
\BIBentryALTinterwordspacing
N.~B. of~Statistics~of China, ``Bulletin of the seventh national census (no. 5),'' 2023. [Online]. Available: \url{https://www.stats.gov.cn/sj/zxfb/202302/t20230203_1901085.html}
\BIBentrySTDinterwordspacing

\bibitem{dementia2023}
\BIBentryALTinterwordspacing
W.~H. Organization, ``\BIBforeignlanguage{en}{Dementia},'' 2023. [Online]. Available: \url{https://www.who.int/news-room/fact-sheets/detail/dementia}
\BIBentrySTDinterwordspacing

\bibitem{elderlyreport2021}
\BIBentryALTinterwordspacing
C.~N.~C. on~Ageings, ``Status and development of care services for the elderly with cognitive impairment,'' 2021. [Online]. Available: \url{https://mp.weixin.qq.com/s/DSCwFE12Dc9lO3GyQopQ7Q}
\BIBentrySTDinterwordspacing

\bibitem{alzheimer_diagnosis_2021}
A.~P. Porsteinsson, R.~S. Isaacson, S.~Knox, M.~N. Sabbagh, and I.~Rubino, ``\BIBforeignlanguage{en}{Diagnosis of {Early} {Alzheimer}’s {Disease}: {Clinical} {Practice} in 2021},'' \emph{\BIBforeignlanguage{en}{The Journal of Prevention of Alzheimer's Disease}}, vol.~8, no.~3, pp. 371--386, Jul. 2021.

\bibitem{venugopalan_multimodal_2021}
J.~Venugopalan, L.~Tong, H.~R. Hassanzadeh, and M.~D. Wang, ``Multimodal deep learning models for early detection of {Alzheimer}’s disease stage,'' \emph{Scientific Reports}, vol.~11, no.~1, p. 3254, Feb. 2021, publisher: Nature Publishing Group.

\bibitem{pan_early_2020}
D.~Pan, A.~Zeng, L.~Jia, Y.~Huang, T.~Frizzell, and X.~Song, ``Early {Detection} of {Alzheimer}’s {Disease} {Using} {Magnetic} {Resonance} {Imaging}: {A} {Novel} {Approach} {Combining} {Convolutional} {Neural} {Networks} and {Ensemble} {Learning},'' \emph{Frontiers in Neuroscience}, vol.~14, May 2020.

\bibitem{reilly2010cognition}
J.~Reilly, A.~D. Rodriguez, M.~Lamy, and J.~Neils-Strunjas, ``Cognition, language, and clinical pathological features of non-alzheimer's dementias: an overview,'' \emph{Journal of communication disorders}, vol.~43, no.~5, pp. 438--452, 2010.

\bibitem{jarrold2014aided}
W.~Jarrold, B.~Peintner, D.~Wilkins, D.~Vergryi, C.~Richey, M.~L. Gorno-Tempini, and J.~Ogar, ``Aided diagnosis of dementia type through computer-based analysis of spontaneous speech,'' in \emph{Proceedings of the Workshop on Computational Linguistics and Clinical Psychology: From Linguistic Signal to Clinical Reality}, 2014, pp. 27--37.

\bibitem{cullen2007review}
B.~Cullen, B.~O’Neill, J.~J. Evans, R.~F. Coen, and B.~A. Lawlor, ``A review of screening tests for cognitive impairment,'' \emph{Journal of Neurology, Neurosurgery \& Psychiatry}, vol.~78, no.~8, pp. 790--799, 2007.

\bibitem{lonie2009screening}
J.~A. Lonie, K.~M. Tierney, and K.~P. Ebmeier, ``Screening for mild cognitive impairment: a systematic review,'' \emph{International Journal of Geriatric Psychiatry: A journal of the psychiatry of late life and allied sciences}, vol.~24, no.~9, pp. 902--915, 2009.

\bibitem{law2012measures}
L.~L. Law, F.~Barnett, M.~K. Yau, and M.~A. Gray, ``Measures of everyday competence in older adults with cognitive impairment: a systematic review,'' \emph{Age and Ageing}, vol.~41, no.~1, pp. 9--16, 2012.

\bibitem{bielak2017cognitive}
A.~A. Bielak, C.~R. Hatt, and M.~Diehl, ``Cognitive performance in adults’ daily lives: Is there a lab-life gap?'' \emph{Research in Human Development}, vol.~14, no.~3, pp. 219--233, 2017.

\bibitem{aggio2018cognition}
N.~M. Aggio, M.~Ducatti, and J.~C. de~Rose, ``Cognition and language in dementia patients: Contributions from behavior analysis,'' \emph{Behavioral Interventions}, vol.~33, no.~3, pp. 322--335, 2018.

\bibitem{kempler2008language}
D.~Kempler and M.~Goral, ``Language and dementia: Neuropsychological aspects,'' \emph{Annual review of applied linguistics}, vol.~28, pp. 73--90, 2008.

\bibitem{bookheimer2002functional}
S.~Bookheimer, ``Functional mri of language: new approaches to understanding the cortical organization of semantic processing,'' \emph{Annual review of neuroscience}, vol.~25, no.~1, pp. 151--188, 2002.

\bibitem{price2012review}
C.~J. Price, ``A review and synthesis of the first 20 years of pet and fmri studies of heard speech, spoken language and reading,'' \emph{Neuroimage}, vol.~62, no.~2, pp. 816--847, 2012.

\bibitem{nasreddine2005montreal}
Z.~S. Nasreddine, N.~A. Phillips, V.~B{\'e}dirian, S.~Charbonneau, V.~Whitehead, I.~Collin, J.~L. Cummings, and H.~Chertkow, ``The montreal cognitive assessment, moca: a brief screening tool for mild cognitive impairment,'' \emph{Journal of the American Geriatrics Society}, vol.~53, no.~4, pp. 695--699, 2005.

\bibitem{della2002empirical}
V.~Della-Maggiore, W.~Chau, P.~R. Peres-Neto, and A.~R. McIntosh, ``An empirical comparison of spm preprocessing parameters to the analysis of fmri data,'' \emph{Neuroimage}, vol.~17, no.~1, pp. 19--28, 2002.

\bibitem{wong2009validity}
A.~Wong, Y.~Y. Xiong, P.~W. Kwan, A.~Y. Chan, W.~W. Lam, K.~Wang, W.~C. Chu, D.~L. Nyenhuis, Z.~Nasreddine, L.~K. Wong \emph{et~al.}, ``The validity, reliability and clinical utility of the hong kong montreal cognitive assessment (hk-moca) in patients with cerebral small vessel disease,'' \emph{Dementia and geriatric cognitive disorders}, vol.~28, no.~1, pp. 81--87, 2009.

\bibitem{buxton2004modeling}
R.~B. Buxton, K.~Uluda{\u{g}}, D.~J. Dubowitz, and T.~T. Liu, ``Modeling the hemodynamic response to brain activation,'' \emph{Neuroimage}, vol.~23, pp. S220--S233, 2004.

\bibitem{bron2015feature}
E.~E. Bron, M.~Smits, W.~J. Niessen, and S.~Klein, ``Feature selection based on the svm weight vector for classification of dementia,'' \emph{IEEE journal of biomedical and health informatics}, vol.~19, no.~5, pp. 1617--1626, 2015.

\bibitem{power2012spurious}
J.~D. Power, K.~A. Barnes, A.~Z. Snyder, B.~L. Schlaggar, and S.~E. Petersen, ``Spurious but systematic correlations in functional connectivity mri networks arise from subject motion,'' \emph{Neuroimage}, vol.~59, no.~3, pp. 2142--2154, 2012.

\bibitem{rajan2021population}
K.~B. Rajan, J.~Weuve, L.~L. Barnes, E.~A. McAninch, R.~S. Wilson, and D.~A. Evans, ``Population estimate of people with clinical alzheimer's disease and mild cognitive impairment in the united states (2020--2060),'' \emph{Alzheimer's \& dementia}, vol.~17, no.~12, pp. 1966--1975, 2021.

\bibitem{woolson2005wilcoxon}
R.~F. Woolson, ``Wilcoxon signed-rank test,'' \emph{Encyclopedia of Biostatistics}, vol.~8, 2005.

\bibitem{lipkin2022probabilistic}
B.~Lipkin, G.~Tuckute, J.~Affourtit, H.~Small, Z.~Mineroff, H.~Kean, O.~Jouravlev, L.~Rakocevic, B.~Pritchett, M.~Siegelman \emph{et~al.}, ``Probabilistic atlas for the language network based on precision fmri data from> 800 individuals,'' \emph{Scientific Data}, vol.~9, no.~1, p. 529, 2022.

\bibitem{raichle2015brain}
M.~E. Raichle, ``The brain's default mode network,'' \emph{Annual review of neuroscience}, vol.~38, pp. 433--447, 2015.

\bibitem{davey2016exploring}
J.~Davey, H.~E. Thompson, G.~Hallam, T.~Karapanagiotidis, C.~Murphy, I.~De~Caso, K.~Krieger-Redwood, B.~C. Bernhardt, J.~Smallwood, and E.~Jefferies, ``Exploring the role of the posterior middle temporal gyrus in semantic cognition: Integration of anterior temporal lobe with executive processes,'' \emph{Neuroimage}, vol. 137, pp. 165--177, 2016.

\bibitem{binder2017current}
J.~R. Binder, ``Current controversies on wernicke’s area and its role in language,'' \emph{Current neurology and neuroscience reports}, vol.~17, pp. 1--10, 2017.

\bibitem{de2007cerebellum}
H.~J. De~Smet, H.~Baillieux, P.~P. De~Deyn, P.~Mari{\"e}n, and P.~Paquier, ``The cerebellum and language: the story so far,'' \emph{Folia Phoniatrica et Logopaedica}, vol.~59, no.~4, pp. 165--170, 2007.

\bibitem{murdoch2010cerebellum}
B.~E. Murdoch, ``The cerebellum and language: historical perspective and review,'' \emph{Cortex}, vol.~46, no.~7, pp. 858--868, 2010.

\bibitem{verma2012semantic}
M.~Verma and R.~J. Howard, ``Semantic memory and language dysfunction in early alzheimer's disease: a review,'' \emph{International journal of geriatric psychiatry}, vol.~27, no.~12, pp. 1209--1217, 2012.

\bibitem{bickel2000syntactic}
C.~Bickel, J.~Pantel, K.~Eysenbach, and J.~Schr{\"o}der, ``Syntactic comprehension deficits in alzheimer's disease,'' \emph{Brain and Language}, vol.~71, no.~3, pp. 432--448, 2000.

\bibitem{martin1983word}
A.~Martin and P.~Fedio, ``Word production and comprehension in alzheimer's disease: The breakdown of semantic knowledge,'' \emph{Brain and language}, vol.~19, no.~1, pp. 124--141, 1983.

\end{thebibliography}

\end{document}